\documentclass{llncs}
\usepackage{graphicx}
\usepackage{amsmath}
\usepackage{url}
\usepackage{hyperref}
\usepackage{bm}
\usepackage{multirow}

\begin{document}
\mainmatter
\title{Pick-and-Learn: Automatic Quality Evaluation for Noisy-Labeled Image Segmentation}
\titlerunning{Quality Evaluation for Noisy-Labeled Image Segmentation}
%

\author{Haidong Zhu \inst{1}\and
Jialin Shi \inst{1}\and
Ji Wu\inst{1,2}}


\authorrunning{H. Zhu et al.}
%
\institute{Department of Electronic Engineering, Tsinghua University, Beijing, China\\
\email{\{zhuhd15,shi-jl16\}@mails.tsinghua.edu.cn}
\and
Institute for Precision Medicine, Tsinghua University, Beijing, China\\
\email{wuji\_ee@mail.tsinghua.edu.cn}}

\maketitle              
\renewcommand{\thefootnote}{}
\footnotetext{H. Zhu and J. Shi contributed equally to this work.}
\begin{abstract}
 Deep learning methods have achieved promising performance in many areas, but they are still struggling with noisy-labeled images during the training process. Considering that the annotation quality indispensably relies on great expertise, the problem is even more crucial in the medical image domain. How to eliminate the disturbance from noisy labels for segmentation tasks without further annotations is still a significant challenge. In this paper, we introduce our label quality evaluation strategy for deep neural networks automatically assessing the quality of each label, which is not explicitly provided, and training on clean-annotated ones. We propose a solution for network automatically evaluating the relative quality of the labels in the training set and using good ones to tune the network parameters. We also design an overfitting control module to let the network maximally learn from the precise annotations during the training process. Experiments on the public biomedical image segmentation dataset have proved the method outperforms baseline methods and retains both high accuracy and good generalization at different noise levels.

\end{abstract}

\section{Introduction}
\begin{figure}
\includegraphics[width=\textwidth]{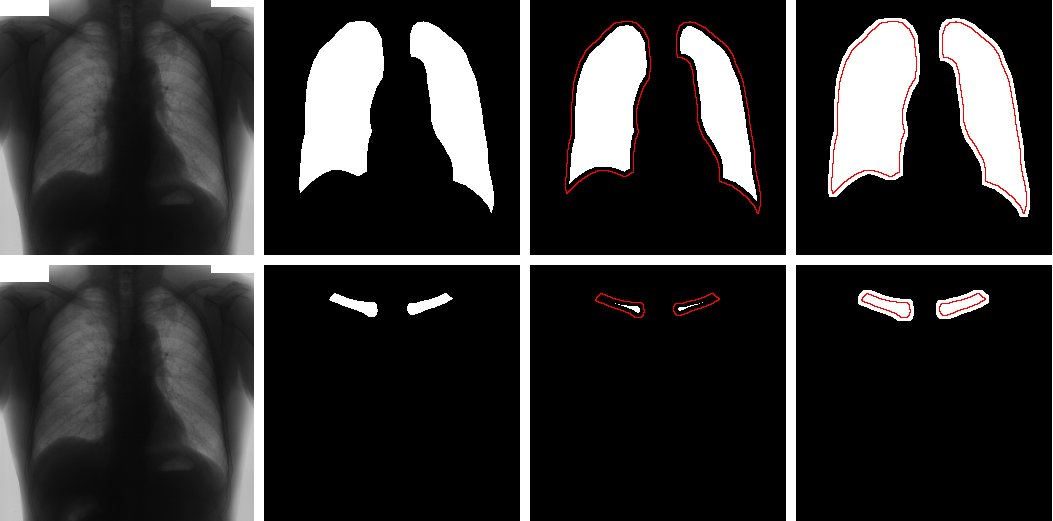}
\caption{Two examples of noisy labels in the segmentation problem. Images in the second row are the clean-annotated ground-truth. The third and fourth columns show two types of noisy labels: dilation and erosion. Correct segmentation boundaries are shown in red.} \label{question}
\end{figure}

Researchers have witnessed the significant improvement of many visual understanding tasks based on deep learning methods \cite{vgg,resnet,unet} in the past few years, but the community is still struggling to get a massive amount of clean and precise labels to train a good model. Noisy-labeled images can seriously damage the performance of the deep neural networks \cite{iclr}. For example, the computer-aided image segmentation for nodes and organs will directly help doctors and surgeons to understand, diagnose and perform the operation. But low-accuracy segmentation model trained on noisy labels, like examples shown in Fig.~\ref{question}, will severely impact the result and can lead to more serious misunderstandings during the surgery. High-quality annotations for the data require the accumulation of years of knowledge and experience, which directly conflicts with the need for the great need of data.

Compared with the solutions for low-quality images, noisy labels are more difficult to deal with if no further annotation for quality is available. Most approaches for low-quality annotations are designed for the classification tasks \cite{tad,nlb,mentornet,jlt,nlcv,tlv,rob}. For example, based on the distribution of the noise, Goldberger et al. \cite{tad} and Patrini et al. \cite{nlb} relied on the relationship between clean data and noisy data to improve the performance. This needs strong reliability of the predicted distribution of the data. Veit et al. \cite{nlcv} focused on tuning the network on clean data before adding noisy data, which needed extra human annotations. Xue et al. \cite{rob} and Jiang et al. \cite{mentornet} showed the improvement by clustering and using the reweighted loss in the final stage, but they ignored the inner-connection between the input images and annotations. In addition, all the methods described above showed improvement in image classification, but none of them can be applied to the segmentation task.

In this paper, we introduce a label quality evaluation strategy to assess the quality of annotations for model training. Based on the conflict between noisy labels and consistency in clean-labeled samples, the method can be applied to segmentation tasks for images with noisy labels. Specifically, we apply this method on biomedical image segmentation task. With the predicted quality score for each sample in the mini-batch, the network re-weights the loss to tune the network. This work makes two main contributions: first, a novel automatic quality evaluation module inspired by the relationship between labels and inputs; and second, an overfitting control module to ensure enough trainable samples and avoid overfitting. Experiments on the public dataset prove the method can retain both high accuracy and good generalization at different noise levels.

\section{Label Quality Evaluation}
\begin{figure}
\includegraphics[width=\textwidth]{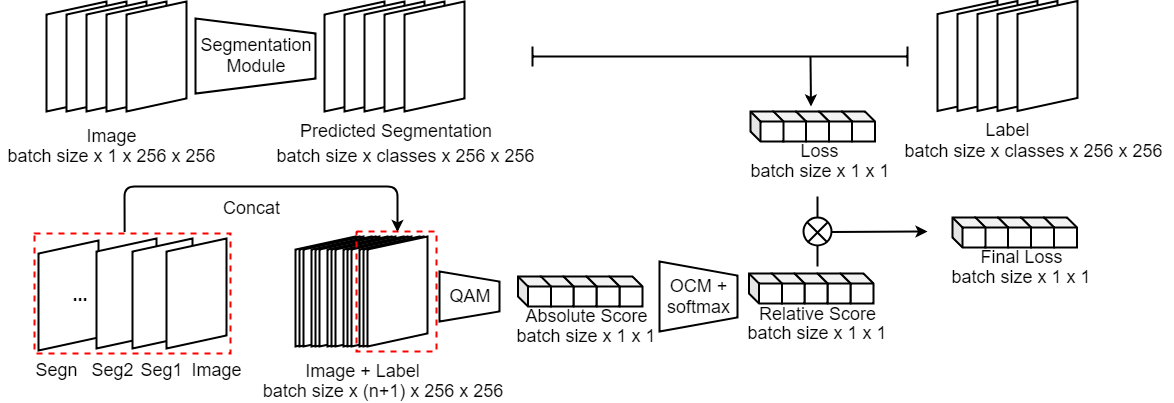}
\caption{The end-to-end architecture of our proposed label quality evaluation strategy. The segmentation module is the CNN structure module for generating segmentation. The quality awareness module (QAM) is a CNN structure network taking the concatenation of the image and its labels, marked as Segn in the image, as input, and running parallelly with the segmentation module. To re-weight the samples in the same mini-batch, the quality awareness module and the overfitting control module (OCM) generate a score for each input annotation, followed by a softmax layer and the multiplication between it and the loss generated by segmentation module. The final loss will be summed up and backpropagated to tune the segmentation and quality awareness modules together.} \label{network}
\end{figure}

Our goal is to use the network to calculate the relative quality score in the same mini-batch and find the conflict information among the noisy labels. With such information, the network can distinguish noisy labels from the good ones. Fig.~\ref{network} illustrates our proposed network structure. Our network is composed of three modules, a feature extraction module, a quality awareness module (QAM) and an overfitting control module (OCM).

\subsection{Quality Awareness Module (QAM)}
One significant modification in our architecture is the quality awareness module, which runs parallelly with the segmentation module to calculate the weight value. During the training process, the predicted segmentation for the noisy-labeled samples might have a higher loss compared with well-annotated ones. Different noise types and noise levels among the noisy-labeled samples can confuse the network and show a slower speed of loss descent.
In this regard, the quality awareness module is proposed to evaluate the potential quality of the labels. During the training process, the quality awareness module supervises the descending speed of the losses in the same mini-batch. After re-weighting the losses, the segmentation module backpropagates based on the re-weighted loss and focuses more on those with the higher weight, while the quality awareness module backpropagates on the same loss.

More specifically, let $x_i$ be the $i^{th}$ input image in the training set and $y_i$ be its annotation. $L_i$ is the $i^{th}$ loss calculated by the segmentation module in the mini-batch with $N$ samples. We use the VGG \cite{vgg} based network $\Theta(\cdot)$ in the experiment and change the number of input channel to $n+1$, where $n$ is the number of segmentation classes, to take both image and labels as input. Also, we replace the final layer as a one-channel average pooling layer to assess the weight for each sample in the mini-batch. A softmax layer is followed to rescale all the score to [0,1]. Instead of naively using the average value of the loss to backpropagate, the final loss is calculated as
\begin{equation}
\begin{split}
    Loss = \sum_{i=1,...,N}\Theta(x_i,y_i)\cdot L_i \\
    s.t. \sum_{i=1,...,N}\Theta(x_i,y_i) = 1, 0 \leq \Theta(x_i,y_i) \leq 1
\end{split}
\end{equation}
to distinguish different quality of the labels.

\subsection{Overfitting Control Module (OCM)}
Although the quality awareness module can separate clean-annotated labels from the noisy ones,  it can cause serious overfitting problems as well, since the loss can further decrease if the quality awareness module has a larger weight on a small subset with clear-labeled samples. Also, if the quality awareness module makes mistakes during the training process, the relative weight of this sample can be too small or too big for the network to correct. In this regard, we add an extra restriction between the quality awareness module and the softmax layer to limit the relative ratio of the output quality score. We use the function
\begin{align}
\Phi(t) = \lambda tanh(t)
\end{align}
as the new score instead of the absolute quality score $t$ produced by quality awareness module.
This will rescale the output score from ($-\infty $,$\infty $) to ($-\lambda$,$\lambda$). After processed by the overfitting control module and the softmax layer, the maximum possible ratio of two relative scores in the same mini-batch decrease from $\infty$ to $e^{2\lambda}$. This ensures that after multiplying the score with the loss, the quality awareness module will not fully ignore any sample in the same mini-batch or ignore clean labels while making mistakes in the early quality evaluation.

\section{Experiments and Results}
In this section, we conducted experiments on a publicly available medical dataset to demonstrate the effectiveness of the label quality evaluation network. We compared our method with the supervised segmentation network without the QAM or OCM modules on different noise levels respectively. We also made a comparison to show the effect of different modules on the performance.

\subsection{Dataset}
We employed a public segmentation medical image dataset JSRT \cite{jsrt} for the experiments. The JSRT dataset contains 247 images and three types of organ structures: heart, clavicles, and lungs. Specifically, each image contains both left and right lungs as well as left and right clavicles. Each image is a 2D grayscale image with 1024x1024 pixels. We randomly split the data into training (165 images) and testing (82 images) subsets in experiments.

\subsection{Implementation Details}
We used the PyTorch in our experiments, and selected the structure of UNet \cite{unet} for the segmentation network. We resized the input image to $256 \times 256$ pixels and set the hyperparameter $\lambda=2$ in the OCM empirically. In order to generate noisy-labeled images with different types and levels of noises, we randomly selected 0\%, 25\%, 50\% and 75\% samples from the training set and further randomly eroded or dilated them with $1\leq n_i\leq 8$ and $5\leq n_i\leq 13$ pixels. We empirically set the learning rate as 0.0001, and the batch size was fixed to 32 in the experiments.

\subsection{Quantitative Results}
In this subsection, we compared our strategies with existing state-of-the-art methods on different levels of noises firstly. Then we conducted ablation experiments to investigate the effectiveness of QAM and OCM. Finally, we evaluated the effectiveness and capabilities of picking clean-labeled samples by using the label quality assessment strategy. In the experiments, we used Dice value
\begin{align}
Dice = \frac{2\times V_{pred}\cap V_{gt}}{V_{pred}+V_{gt}}
\end{align}
as our evaluation criteria, where $V_{pred}$ represents the segmentation area of prediction and $V_{gt}$ is the area of annotation.

\begin{table}
\centering
\caption{Results on JSRT dataset}\label{tab1}
\begin{tabular}{ p{70pt} p{50pt} p{60pt} p{38pt} p{38pt} p{38pt} p{33pt}}
\hline
Noise percentage & Noise level &  Strategy & Lungs & Heart & Clavicles & Average\\
\hline
No noise & - &  baseline & 0.943 & 0.941 & 0.862 & 0.915\\
No noise & - &  QAM & 0.939 & 0.923 & 0.831 & 0.898\\
No noise & - &  QAM+OCM & 0.941 & 0.940 & 0.852 & 0.911\\
\hline
25\% noise & $1\leq n_i\leq 8$ &  baseline & 0.868 & 0.888 & 0.538 & 0.765\\
25\% noise & $1\leq n_i\leq 8$ &  QAM & 0.925 & 0.926 & 0.748 & 0.866\\
25\% noise & $1\leq n_i\leq 8$ &  QAM+OCM & 0.936 & 0.925 & 0.823 & 0.895\\
\hline
50\% noise & $1\leq n_i\leq 8$ &  baseline & 0.873 & 0.884 & 0.539 & 0.765\\
50\% noise & $1\leq n_i\leq 8$ &  QAM &0.922 & 0.925 & 0.726 & 0.857\\ 
50\% noise & $1\leq n_i\leq 8$ &  QAM+OCM & 0.936 & 0.929  & 0.828 & 0.898\\
\hline
75\% noise & $1\leq n_i\leq 8$ &  baseline& 0.820 & 0.828 & 0.512 & 0.720\\
75\% noise & $1\leq n_i\leq 8$ &  QAM & 0.898 & 0.825 & 0.536 & 0.753\\
75\% noise & $1\leq n_i\leq 8$ &  QAM+OCM & 0.937 & 0.939  & 0.809 & 0.895\\
\hline
25\% noise & $5\leq n_i\leq 13$ &  baseline & 0.865 & 0.857 & 0.422 & 0.715\\
25\% noise & $5\leq n_i\leq 13$ &  QAM & 0.893 & 0.835 & 0.615 & 0.781\\
25\% noise & $5\leq n_i\leq 13$ &  QAM+OCM & 0.935 & 0.935 & 0.801 & 0.890\\
\hline
50\% noise & $5\leq n_i\leq 13$ & baseline & 0.755 & 0.807 & 0.393 & 0.652\\
50\% noise & $5\leq n_i\leq 13$ &  QAM & 0.828 & 0.853 & 0.491 & 0.714\\
50\% noise & $5\leq n_i\leq 13$ &  QAM+OCM & 0.942 & 0.942 & 0.853 & 0.912\\
\hline
75\% noise & $5\leq n_i\leq 13$ & baseline & 0.745 & 0.738 & 0.381 & 0.621\\
75\% noise & $5\leq n_i\leq 13$ &  QAM & 0.770 & 0.772 & 0.366 & 0.636\\
75\% noise & $5\leq n_i\leq 13$ &  QAM+OCM & 0.938 & 0.937 & 0.801 & 0.892\\
\hline
\end{tabular}
\end{table}

\subsubsection{Comparison with baseline methods}  
\begin{figure}[h]
\centering
\begin{minipage}[c]{0.5\textwidth}
\centering
\includegraphics[width=\textwidth]{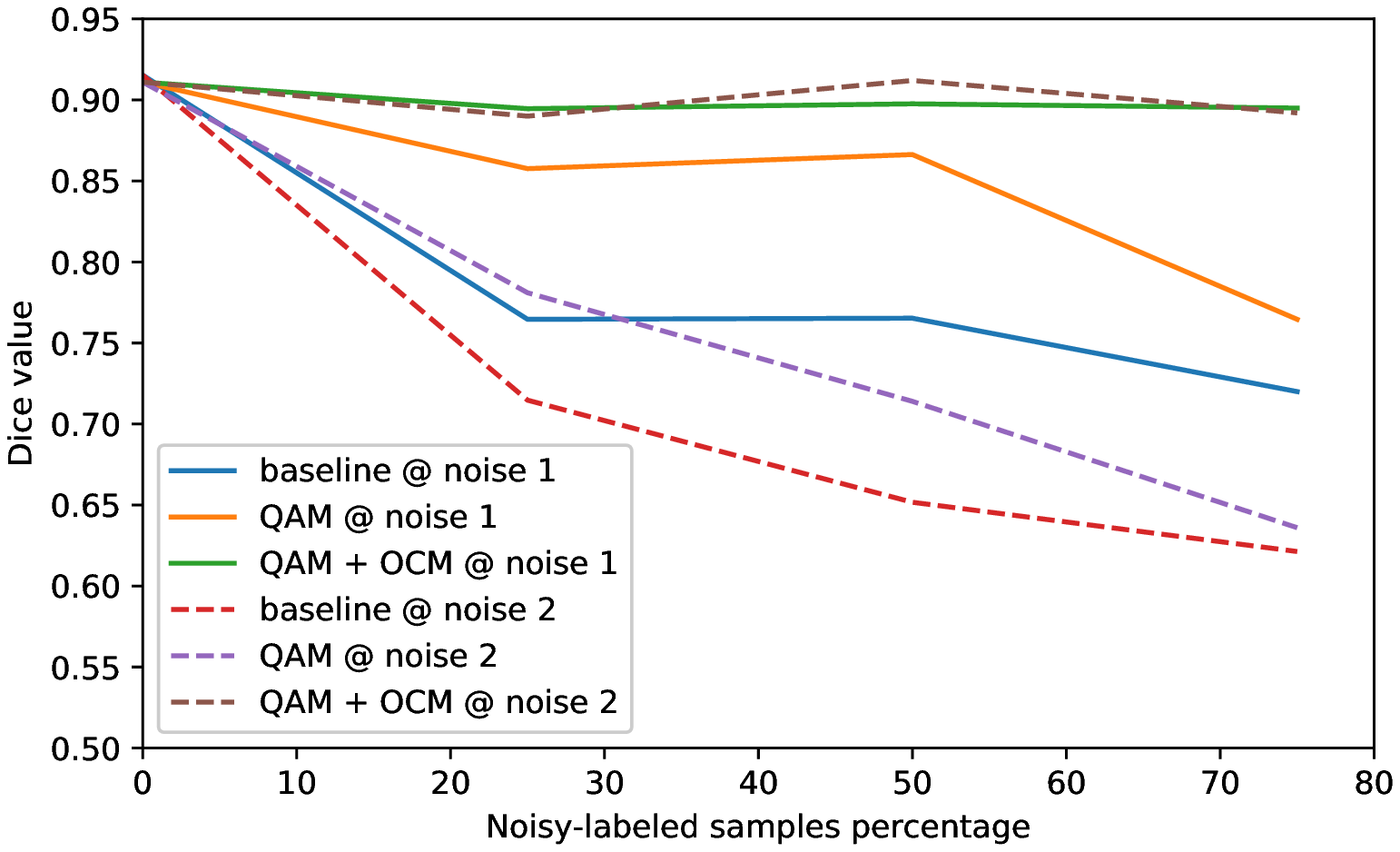}
\end{minipage}%
\begin{minipage}[c]{0.5\textwidth}
\centering
\includegraphics[width=\textwidth]{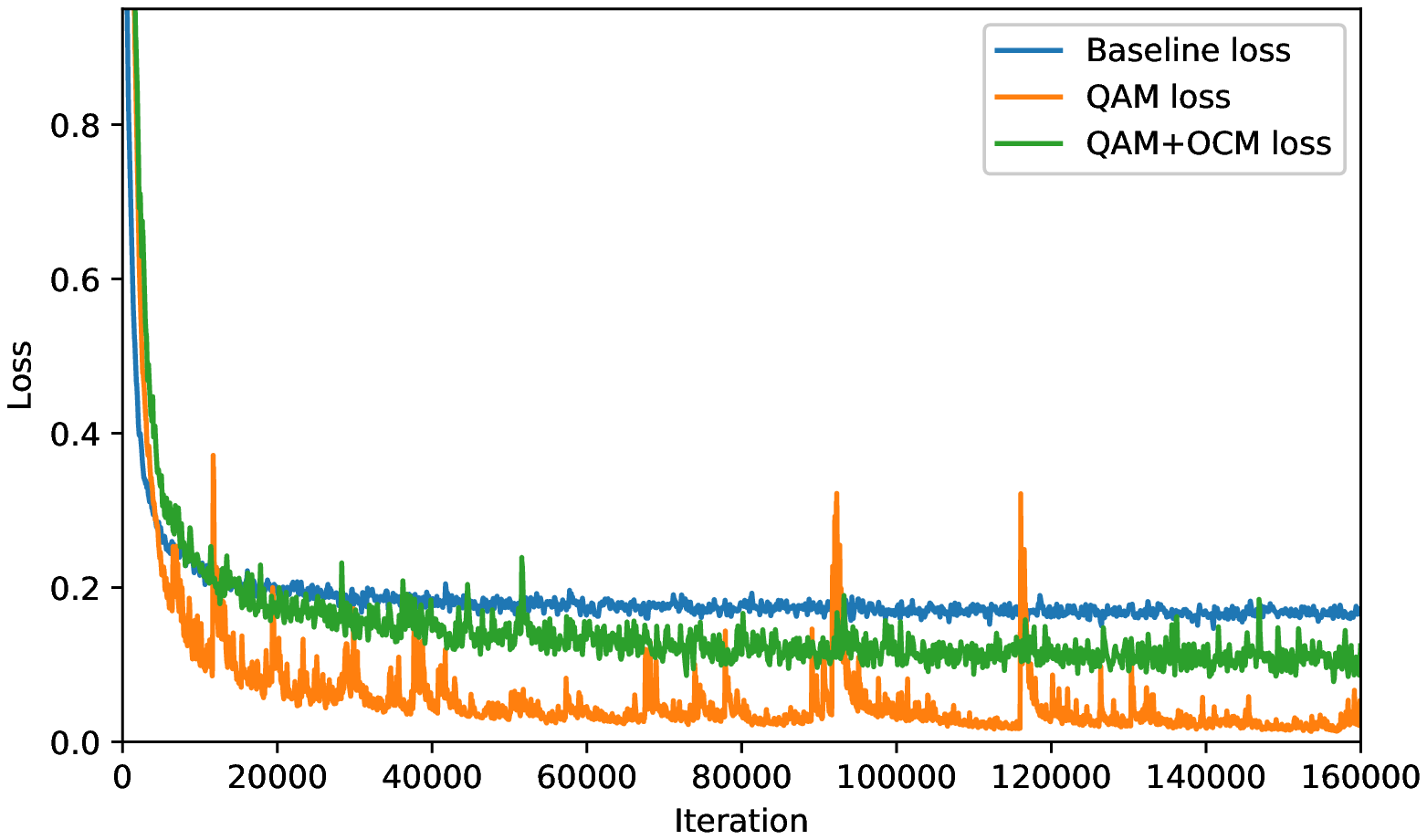}
\end{minipage}

\caption{Average class accuracy and loss plots of different noise levels on JSRT. Noise 1 and noise 2 represent $1 \leq n_i \leq 8$ and $5 \leq n_i \leq 13$ respectively. Loss curves belong to models trained on the training set with 50\% of labels dilated or eroded 8 to 13 pixels.}
\label{curve}
\end{figure}

We conducted the experiments on the datasets described in the previous section. We trained the network on the training set with different levels of noisy labels and tested on clean labels. Table~\ref{tab1} illustrates the experimental results with and without the label quality evaluation strategy. On clean-annotated dataset, both these methods have high accuracy in all three segmentation tasks. But as the noise level increases both in the area and ratio, the segmentation performance for the baseline method decreases sharply. The results for smaller anatomical structures, such as clavicles, suffer terribly.  
However, when adding the modules of quality awareness and overfitting control, the segmentation network can recover and even retain its high accuracy. We observe that the performance of our method is comparable to the model trained on clean datasets. Fig.~\ref{curve} shows the average class accuracy and loss curves of the network on the noisy-labeled datasets with different noise levels. Compared with high loss values without adding the modules, the quality awareness network can find out the relative better examples and show lower loss during training. 

\subsubsection{Effect of overfitting control module} 
We also evaluated the influence of the overfitting control module in the experiment. The quality awareness module without overfitting control shows smaller training loss on the training set on all noise levels, but the test score is below the result of the experiments with the overfitting control module. When noise level increases, the final test accuracy for segmentation network is relatively high at some noise levels, but it performs badly and is extremely unstable. For example, for clavicles, when 75\% of the labels are noisy-labeled, and $5 \leq n_i \leq 13$, the segmentation accuracy without the overfitting control module decreases from 0.862 to 0.366, which is even worse compared with baseline method, but with such strategy, the network can still keep a 0.801 segmentation accuracy. The network with the overfitting control module consistently retains its high accuracy when the noise level increases and shows a stable and competitive result. Experiment results demonstrate that without overfitting control module, the network suffers severe overfitting and will lose its ability in generalization.

\subsubsection{Evaluating the selecting policy} 
\begin{figure}[h]
\centering
\begin{minipage}[c]{0.5\textwidth}
\centering
\includegraphics[width=\textwidth]{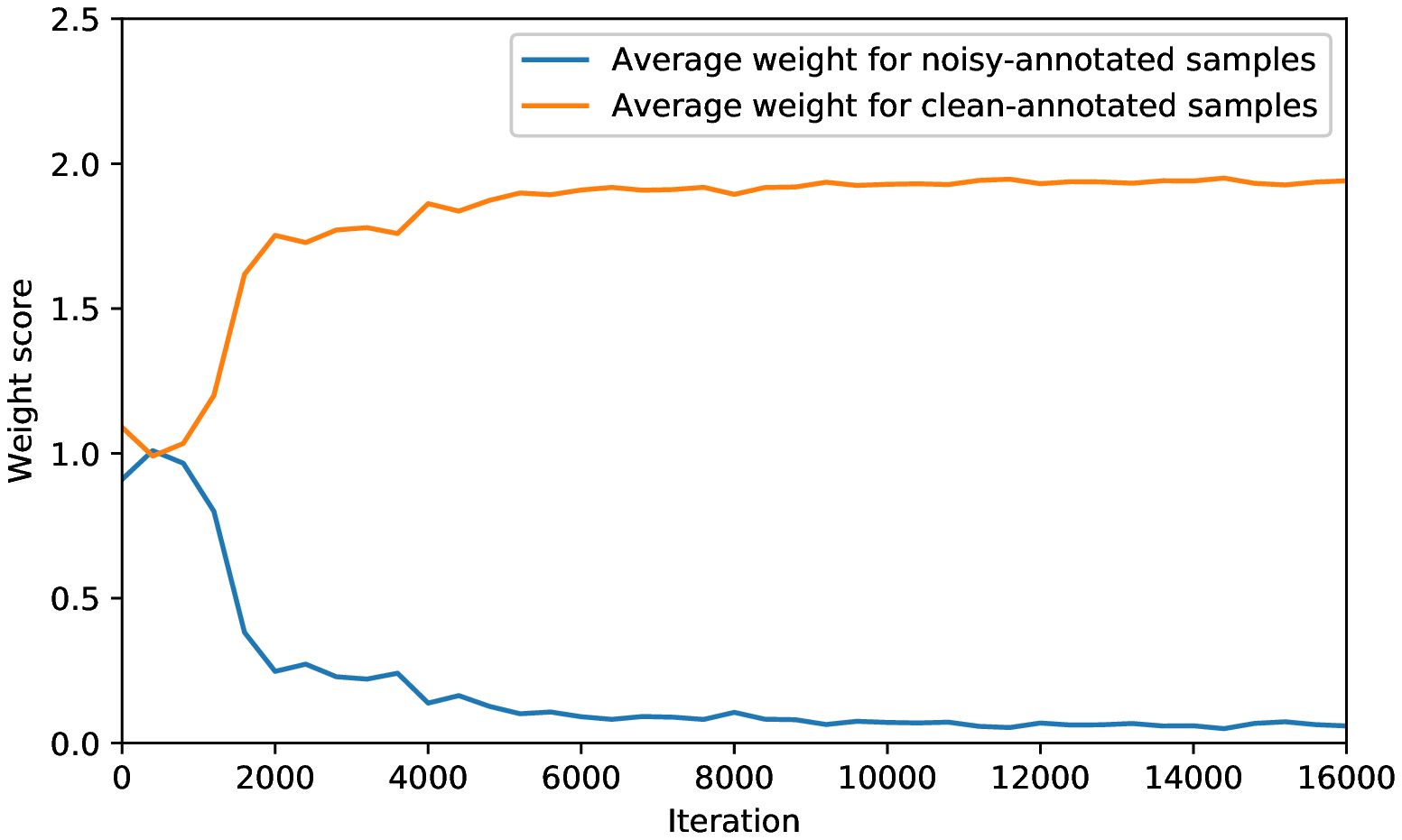}
\end{minipage}%
\begin{minipage}[c]{0.5\textwidth}
\centering
\includegraphics[width=\textwidth]{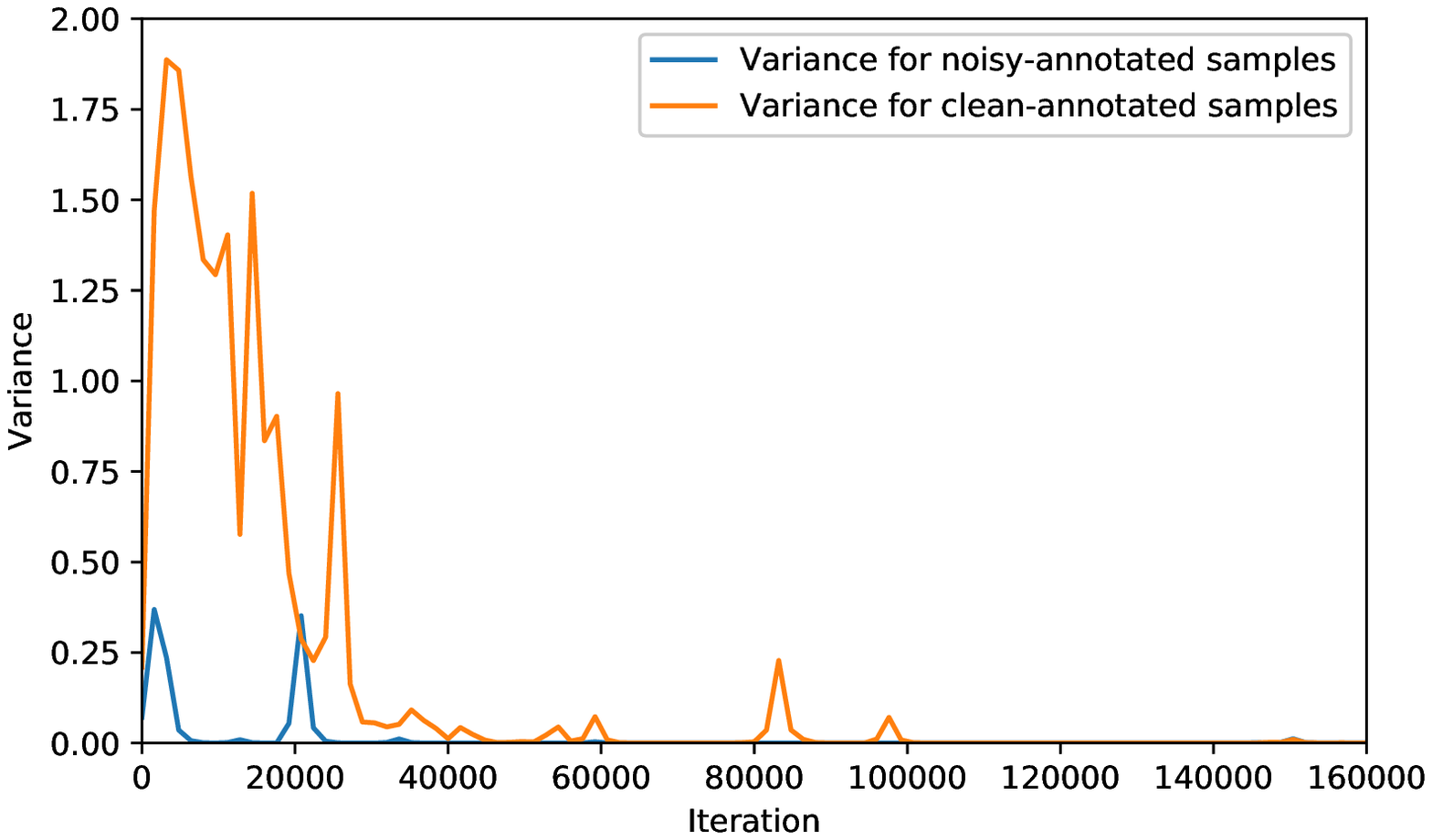}
\end{minipage}
\caption{Relative weights and variances for clean and noisy-labeled data.} \label{select}
\end{figure}

We evaluated the performance of the label quality evaluation strategy during the training process. At different steps, we calculated the average weight of the noisy-labeled data and clean data. We tested the network on the dataset with 50\% of its mini-batch are $5 \leq n_i \leq 13$ pixels dilated or eroded from its correct annotation. Fig.~\ref{select} shows the analysis. 
At the beginning of the training, the network cannot separate between these two types of data. However, the relative score given to the clean samples gradually goes higher, and the ratio between good samples and noisy-labeled samples got larger, showing that quality awareness module can pick out some of the clean-labeled samples first. As training goes further, the variance value of the weight decreases, indicating that quality evaluation strategy can gradually find a consistent criteria of picking clean annotated samples and gradually increase its recall rate. The network can find out bad samples very quickly, and with the overfitting control module, it can gradually find more clean annotated samples for training. The weight value for the samples with same quality is getting more and more stable. 

\section{Conclution}
In this paper, we have proposed a method to tune the segmentation network on noisy-labeled datasets called label quality evaluation strategy, which consists of three parts: segmentation module, quality awareness module, and overfitting control module. Quality awareness module can evaluate the relative quality of the labels in the mini-batch and re-weight them, and the overfitting control module can retain the generalization of the network. Compared with models without this method, our quality awareness model keeps high segmentation accuracy when the noise level increases. We have presented and analyzed the efficiency and necessity of these two modules. Experimental results on the benchmark dataset demonstrate that our quality awareness network outperforms other methods on segmentation tasks and achieves very competitive results on the noisy-labels dataset with the state-of-the-art supervised methods trained on clean-annotated data.

\subsubsection{Acknowledgement} This work is supported by the National Key Research and Development Program of China (No.2018YFC0116800).

%
%
%
%

\end{document}